\documentclass{article} % For LaTeX2e
\usepackage{iclr2021_conference,times}

\usepackage{amsmath,amsfonts,bm}

\def\eqref#1{equation~\ref{#1}}
\def\1{\bm{1}}

\DeclareMathAlphabet{\mathsfit}{\encodingdefault}{\sfdefault}{m}{sl}
\SetMathAlphabet{\mathsfit}{bold}{\encodingdefault}{\sfdefault}{bx}{n}

\usepackage{hyperref}
\usepackage{url}

\title{Representative \& Fair Synthetic Data}

\author{Paul Tiwald \\
MOSTLY AI \\
Vienna, Austria \\
\texttt{paul.tiwald@mostly.ai} \\
\And
Alexandra Ebert \\
MOSTLY AI \\
Vienna, Austria \\
\texttt{alexandra.ebert@mostly.ai} \\
\And
Daniel T. Soukup \\
Toronto, ON, Canada \\
\texttt{daniel.t.soukup@gmail.com} \\
}

\usepackage{graphicx}
\graphicspath{ {.} }

\iclrfinalcopy

\begin{document}

\maketitle

\begin{abstract}
Algorithms learn rules and associations based on the training data that they are exposed to. Yet, the very same data that teaches machines to understand and predict the world, contains societal and historic biases, resulting in biased algorithms with the risk of further amplifying these once put into use for decision support. Synthetic data, on the other hand, emerges with the promise to provide an unlimited amount of representative, realistic training samples, that can be shared further without disclosing the privacy of individual subjects. We present a framework to incorporate fairness constraints into the self-supervised learning process, that allows to then simulate an unlimited amount of representative as well as fair synthetic data. This framework provides a handle to govern and control for privacy as well as for bias within AI at its very source: the training data. We demonstrate the proposed approach by amending an existing generative model architecture and generating a representative as well as fair version of the UCI Adult census data set. While the relationships between attributes are faithfully retained, the gender and racial biases inherent in the original data are controlled for. This is further validated by comparing propensity scores of downstream predictive models that are trained on the original data versus the fair synthetic data. We consider representative \& fair synthetic data a promising future building block to teach algorithms not on historic worlds, but rather on the worlds that we strive to live in.
\end{abstract}

\section{Introduction}\label{sec:intro}

Decision processes within public as well as private organizations are increasingly supported by machine learning algorithms. These come with the benefit of being fast, scalable, consistent, and oftentimes more accurate as well as objective\footnote{E.g., \cite{gates2002automated} showed how the adoption of automated underwriting in mortgage lending contributed to the increase of approval rates for minority and low-income applicants by 30\% while improving the overall accuracy of default predictions.} compared to human judgement. However, these algorithms depend on the provided training data to represent, understand and predict the world. And thus, any historic discrimination against groups based on their gender, race or sexual orientation, that is already present in the training data, influences the model as well \citep{kearns2019ethical, barocas2017fairness}. If not addressed, these systemic biases end up in data sets that decision-making algorithms are trained on. Subsequently, the biased algorithms make unfair decisions, perpetuating, and actually amplifying the biases in our society. More often than not, engineers or managers are not aware of the issue at hand, making unwanted bias one of the key challenges for establishing Ethical AI.

AI-generated synthetic data emerges as a novel approach to safely share (training) data with third parties. As it strives to retain statistical information of an original data set, it does break the 1:1 relationship to actual records and thus can effectively prevent the re-identification of individuals. But the very same generative models, that are tasked to yield realistic, representative synthetic samples, can be amended to shift patterns and relationships in order to make the data also fair. For simplicity, we start out with adopting the concept of statistical parity for the purpose of this introduction, whereas the approach can also be expanded to other fairness definitions. Statistical parity requires that a classifier gives equal probability with respect to a target variable, independent of whether a subject is or is not contained within a protected group.

There are three points in the machine learning life cycle where one can mitigate bias: at the source, by changing the training data; during the modeling phase by using additional fairness constraints; and as a post-processing step, by revising the algorithm’s decisions in favor of a sensitive group. Naive data-level techniques, such as oversampling methods, have the risk of skewing important data distributions when mitigating imbalances. Similarly, simply excluding the sensitive column, like race or gender, can have detrimental effects \citep{besse2020survey}, as the effect can still be present, yet less apparent, via proxy attributes.\footnote{Imagine we know which neighborhood a person lives in, the brand and model of the person’s mobile phone, the car this person drives, where this person buys her/his clothes, etc. Given some of the above information, we humans can make an educated guess on this person’s sex, skin color, and other attributes. And since algorithms are better in analyzing patterns like this, they will definitely detect these correlations and exploit them, leading again to unfair predictions and decisions. We could actually go one step further and say that leaving the “sex” column in the data set is better for fairness because it offers a clear handle to enforce fairness constraints such as statistical parity.} Thus, it requires a complete representation of the data structure to control for proxy variables in order to properly account for fairness. And for that reason, we propose to incorporate fairness constraints into the training data itself to govern for AI bias already at its source.

%One of the first ideas to try when creating a fair data set for machine learning is to drop the sensitive column. In the presented case that’s the “sex” attribute. At first sight, this sounds like a good and easy-to-implement solution but, unfortunately, it can actually cause more harm than good. On one hand, what makes this approach fail can be so-called proxy or hidden proxy columns. Imagine we know which neighborhood a person lives in, the brand and model of the person’s mobile phone, the car this person drives, where this person buys her/his clothes, etc. Given some of the above information, we humans can make a pretty educated guess on this person’s sex, skin color, and other attributes. And since algorithms are better in analyzing patterns like this, they will definitely detect these correlations and exploit them, leading again to unfair predictions and decisions. We could actually go one step further and say that leaving the “sex” column in the data set is better for fairness because it offers a clear handle to enforce fairness constraints such as statistical parity. To give another example from criminal justice,  women on average are less likely to commit future violent crimes than men with similar criminal records. So, a gender-neutral assessment can overestimate a woman’s recidivism risk. 

%\section{Concept}

Generative deep neural networks are typically tasked to optimize an accuracy loss, in order to yield a model that provides truly new, synthetic samples, that are representative of the original records. That loss measures how well the fitted function matches the observed distributions of the real data. In order to get representative \& fair data, we propose to add an additional fairness loss to the loss function, which penalizes any violation of the statistical parity. That is, we optimize for a combined loss: a weighted sum of the accuracy loss and a fairness loss where the fairness loss is proportional to the deviation from the empirically estimated statistical parity. With all other things equal, this approach allows to explicitly trade off accuracy with fairness, by shifting weights from one loss component to the other.\footnote{A very similar approach for training fair classifiers is described in \cite{manisha2018fnnc}, with an implementation available at \href{https://github.com/yoshavit/fairml-farm}{https://github.com/yoshavit/fairml-farm}.}

%Let’s consider a population that can be split into groups by a sensitive attribute S, such as gender, skin color, age or any other property. Then consider another target attribute T that contains sensitive information on the population such as income, whether or not people spent time in prison or credit history.

%Using this approach, we successfully removed the income inequality with respect to gender from the synthetic version of the Adult data set. We did this with very little compromise on other aspects of data accuracy: for example, you can see we preserved the original Male/Female ratio perfectly.

\section{Empirical Demonstration}\label{sec:results}

\subsection{Impact on Descriptive Statistics}

\begin{figure}[ht]
\begin{center}
\includegraphics[width=13cm]{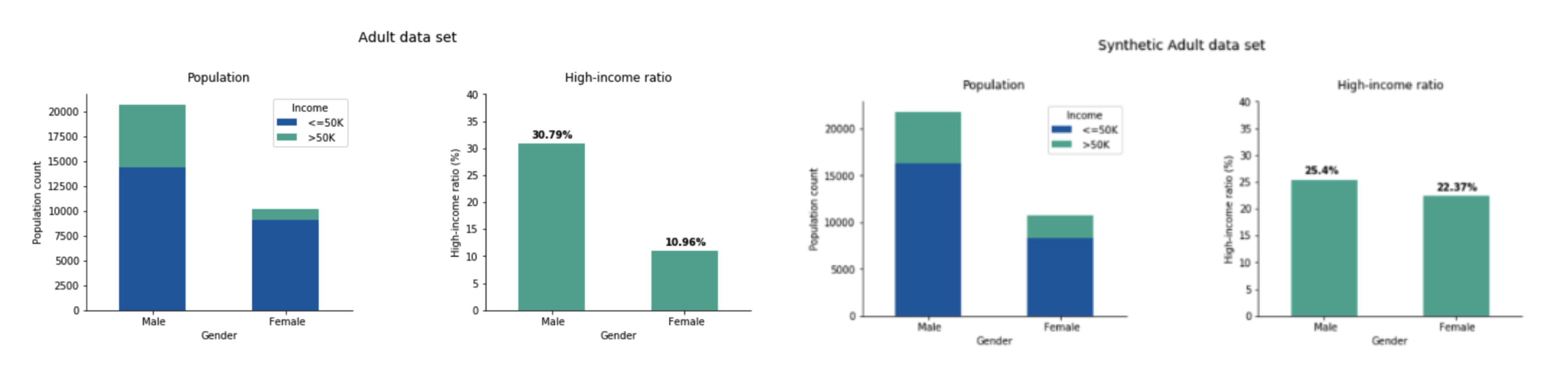}
\end{center}
\caption{The income gap is significantly mitigated in the synthetic data.}\label{fig:adult_gender}
\end{figure}

In order to demonstrate the effectiveness of the concept, we expanded an existing data synthesizer for tabular data with the additional loss component for fairness. We chose to synthesize the well-known Adult income data set \citep{UCIrepo} end-to-end 50 times, each time training the generative model with a parity fairness constraint and treating gender as the protected attribute. Whereas the original data set exhibits a clear gender imbalance within the high-income class, that bias is successfully mitigated in the synthetic data, as is reported in Figure \ref{fig:adult_gender}. With regards to parity, it is common to compare not just the difference but the fraction of high-income male ratio to high-income female ratio. This fraction is called the disparate impact and it is an industry-standard to ask for at least 0.8, the so-called four-fifth rule. In the original data set, this fraction is roughly 10/30 = 0.33, a quite severe disparate impact violation but the bias-corrected synthetic data is at 22/25 = 0.88, well over the threshold.

%Actually, we repeated the whole process 50 times and the plotted numbers are the average ratios over these independent runs. The income-ratios slightly varied across the 50 experiments but this variance (rooted in the stochastic nature of our training and generation process) was quite small: 1.2\% and 1.3\% for the Male and Female ratios, respectively. As apparent from the plots, the synthetization corrected the income gap: 25\% of the synthetic males are high earners (instead of the real 30\%) and 22\% of the synthetic females are high earners (while the original value was 11\% only). 

\begin{figure}[ht]
\begin{center}
\includegraphics[width=13cm]{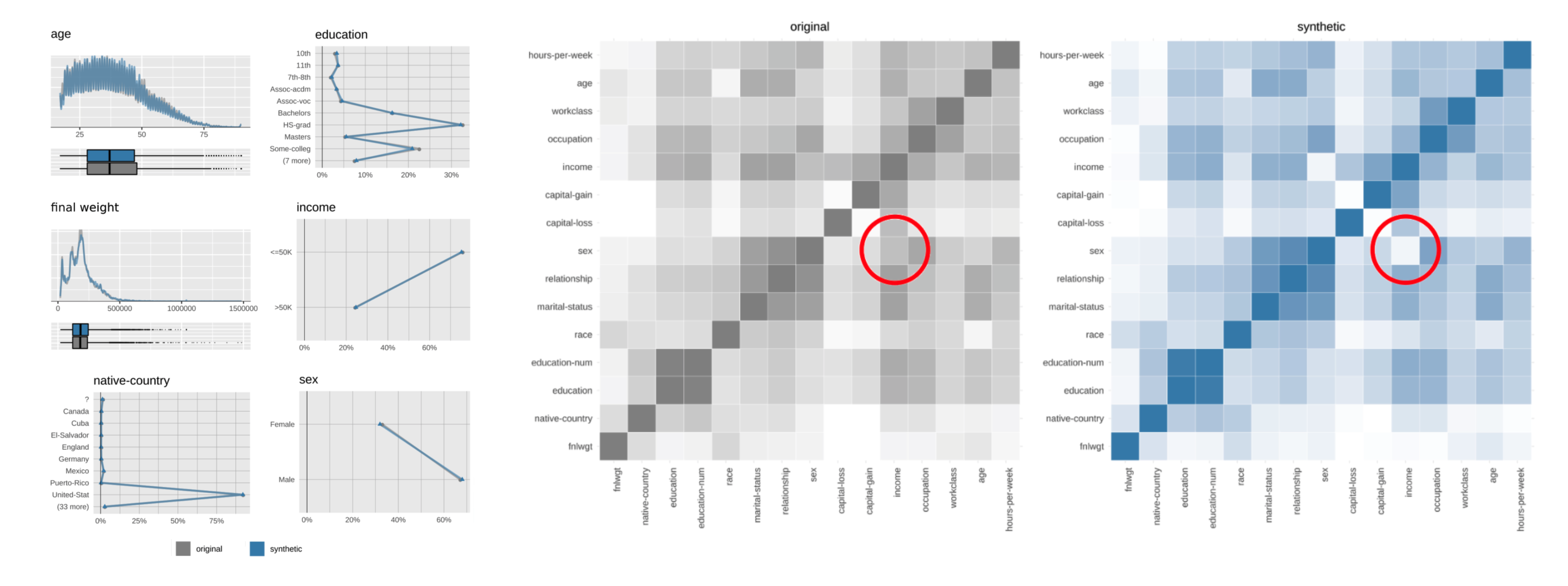}
\end{center}
\caption{Comparison of univariate and bivariate statistics for original and synthetic data.}\label{fig:adult_descriptive}
\end{figure}

Equally important, the additional parity constraint during model training did not diminish the quality of the synthetic data itself (see Figure \ref{fig:adult_descriptive}). The univariate distributions of the original data remain perfectly preserved, including the population-wide male-to-female ratio, as well as the high-earner-to-low-earner ratios. Also, the bivariate correlations are in good agreement with the original data for all but one relationship. Given the statistical parity constraint “income must not depend on sex”, these two attributes should not be correlated. While in the original data, there is a clear “sex”-”income” correlation, this dependency is almost reduced to noise level in the representative \& fair synthetic data set.

\begin{figure}[ht]
\begin{center}
\includegraphics[width=13cm]{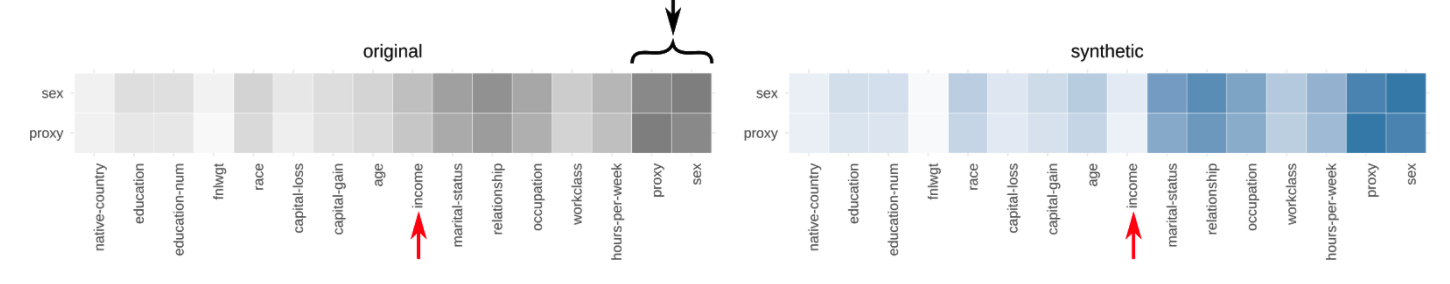}
\end{center}
\caption{The parity-fairness constraint also accounts for proxy attributes for gender.}\label{fig:adult_proxy_fixed}
\end{figure}

We conducted further experiments to study the impact of proxy attributes, and to see whether they can still introduce unfairness through a backdoor, as they are not explicitly accounted for in the parity constraint. We added an artificial feature to the Adult data set named “proxy”, that is strongly correlated with gender. For females, “proxy” equals to 1 in 90\% of all cases and equals to 0 for the remaining 10\%. For males, the percentages are swapped. A subset of the resulting correlation plots is displayed in Figure \ref{fig:adult_proxy_fixed}. The original data shows strong correlation between “sex” and “proxy”, as well as "sex"/"proxy" and “income”. For the synthetic data, the correlation between “sex” and “proxy” remains intact, whereas the correlation of both of these to “income” is significantly reduced due to the induced fairness constraint. This demonstrates that the parity also successfully accounts for (hidden) proxy attributes.

\begin{figure}[ht]
\begin{center}
\includegraphics[width=10cm]{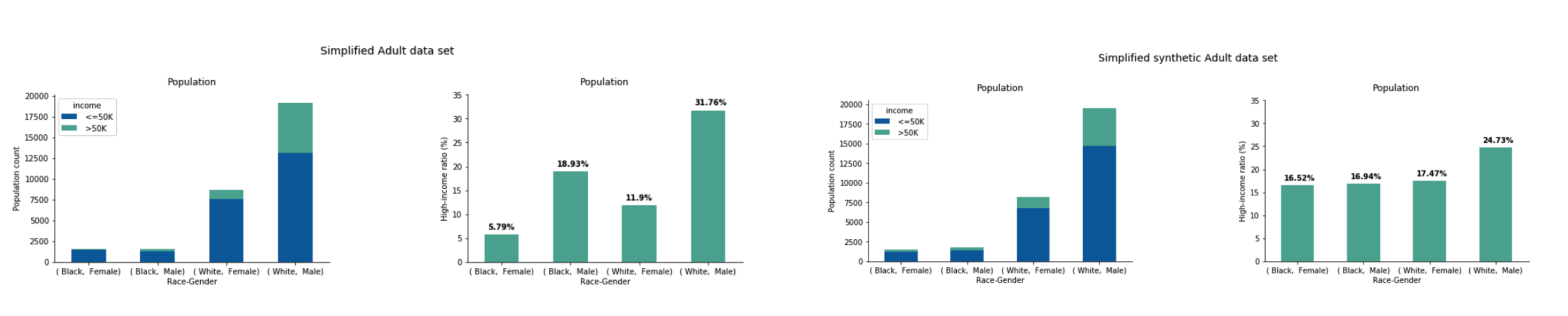}
\end{center}
\caption{Mitigating bias with respect to gender and race simultaneously.}\label{fig:adult_race_gender_fixed}
\end{figure}

In the Adult data set, gender is not the only sensitive attribute: if the generative model is trained incorporating a fairness constraint for “race”, we can achieve similar results as reported for gender. But what if we want to account for fairness with respect to multiple attributes at the same time? In this case, one must be careful what ratios to optimize: if we were to simply put fairness losses independently on race and gender then the algorithm might fall into the mistake of “fairness gerrymandering” \citep{kearns2018preventing}. That is, the new data set would look fair with respect to both gender and race individually, but we would see high imbalances when restricted to gender and race simultaneously. Taking this into account, our proposed approach is capable to yield synthetic data with significantly balanced high-income ratios across all four groups given by race and gender (see Figure \ref{fig:adult_race_gender_fixed}). It is apparent that we did not achieve complete parity but this difference can be further lowered by giving higher weight to the fairness loss against the accuracy loss.

\subsection{Impact on Downstream Models}

We further investigated whether models trained on the de-biased data set yield more fair algorithms. To this end, we fit logistic regression classifiers to predict the income level using all other attributes as predictors. The models were fitted both on the original and the bias-corrected synthetic data and then both were tested on a holdout from the original data. Moreover, we repeated the model training procedure 50 times with independently generated synthetic data. Table \ref{tab:auc} reports the mean performance of the real and synthetic models over these experiments. The synthetically fitted models have very competitive performance and generalize well to the unseen real data.

\begin{table}[ht]
\centering
\begin{tabular}{llll}
          & \textbf{Accuracy}           & \textbf{AUC}                & \textbf{F1-Score}            \\
\hline
original  & \multicolumn{1}{c}{85.49\%} & \multicolumn{1}{c}{91.17\%} & \multicolumn{1}{c}{67.04\%}  \\
synthetic & \multicolumn{1}{c}{84.13\%} & \multicolumn{1}{c}{88.34\%} & \multicolumn{1}{c}{59.8\%}   \\
          &                             &                             &                             
\end{tabular}
\caption{Average predictive performance of downstream models for income.}
\label{tab:auc}
\end{table}

\begin{figure}[ht]
\begin{center}
\includegraphics[width=10cm]{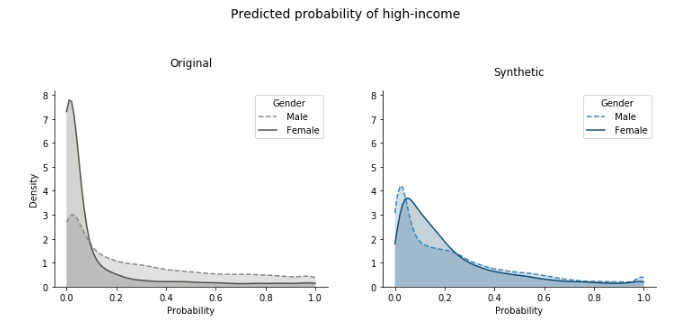}
\end{center}
\caption{Propensity distribution of downstream models for income.}\label{fig:adult_pred_prob}
\end{figure}

Moreover, the models trained on the synthetic data treat the classes of the sensitive attribute (gender, in this case) nearly equally. Figure \ref{fig:adult_pred_prob} visualizes the propensity scores of the corresponding predictive models. The model fitted on the original data assigns a significantly lower probability to women being in the high-income class when compared to male subjects. However, for the model fitted on synthetic data that discrepancy in score assignment is reduced, and the gender-specific distributions largely align. This is exactly the group fairness that parity is designed to capture. The important thing to keep in mind though is that the predictive-model training itself did not involve any type of optimization to fairness and the evaluation is also on the biased original data. So this fair outcome is solely due to using bias-corrected synthetic data for the training. Our results align with the findings of research conducted at Carnegie Mellon University into fair representations of data \citep{zhao2019inherent}. We see that our fairness-constrained synthetic data solution learns to represent data points in a way that removes the dependencies between the sensitive and target attribute while preserving other relationships.

%\section{Discussions}\label{sec:discussions}

%In summary, the introduction of (parity) fairness to our software solution shows very promising results. The quality and accuracy of the synthetic data remain high, the privacy of data subjects is protected, and parity-fairness is guaranteed. All these properties make private and fair synthetic data readily available for further application.

%The notion of fairness (in particular, statistical parity) and synthetic data go together very well. Not only can we generate highly accurate synthetic data but we can also steer the generation to almost perfectly mitigate strong biases in the original data sets. The additional fairness constraint in the training loss of our generative models fine-tunes the correlation structure between attributes such that these biases are strongly reduced. Privacy and (parity) fairness are further preserved in downstream tasks: an out-of-the-box classifier model when trained on fair synthetic data makes fair predictions even on biased input.

%There are many inherent risks in automated decision making and in the use of data sets that do not reflect the world we strive to live in. Historical and measurement biases skew predictive models which in turn affect millions of people who are applying for loans or submitting job applications. As data scientists, engineers, and business leaders, we are responsible to address these issues as best as we can.

\bibliography{iclr2021_conference}
\bibliographystyle{iclr2021_conference}

\appendix
\section{Appendix}

NOTE TO REVIEWERS: For camera-ready version we shall provide a link to download the representative \& fair datasets from the empirical section.

\end{document}